\documentclass{article}
\usepackage{spconf,amsmath,graphicx,hyperref}
\usepackage[T1]{fontenc}
\usepackage[utf8]{inputenc}

\usepackage{booktabs}
\usepackage{graphicx}
\usepackage{subfigure}
\usepackage{algorithm}
\usepackage{algorithmic}
\usepackage{balance}
\usepackage{xcolor}
\usepackage{float}
\usepackage{amssymb}
\usepackage{multirow}  % Make sure this is in your preamble
\usepackage{bbm}
\usepackage{tikz}

\title{AirGlove: Exploring Egocentric 3D Hand Tracking and Appearance Generalization for Sensing Gloves}
 \name{Wenhui Cui$^{1, 2}$\sthanks{Work done during internship at Meta.}, Ziyi Kou$^1$, Chuan Qin$^1$, Ergys Ristani$^1$, Li Guan$^1$}
\address{$^1$ Meta
        Reality Labs Research \\
        $^2$  Ming Hsieh Department of Electrical and Computer Engineering, University of Southern California}
\begin{document}
%\ninept
%
\maketitle
\begin{abstract}

Sensing gloves have become important tools for teleoperation and robotic policy learning as they are able to provide rich signals like speed, acceleration and tactile feedback. A common approach to track gloved hands is to directly use the sensor signals (e.g., angular velocity, gravity orientation) to estimate 3D hand poses. However, sensor-based tracking can be restrictive in practice as the accuracy is often impacted by sensor signal and calibration quality. 
Recent advances in vision-based approaches have achieved strong performance on human hands via large-scale pre-training, but their performance on gloved hands with distinct visual appearances remains underexplored. 
In this work, we present the first systematic evaluation of vision-based hand tracking models on gloved hands under both zero-shot and fine-tuning setups.
Our analysis shows that existing bare-hand models suffer from substantial performance degradation on sensing gloves due to large appearance gap between bare-hand and glove designs. 
We therefore propose \emph{AirGlove}, which leverages existing gloves to generalize the learned glove representations towards new gloves with limited data.
% To address this limitation, we propose \emph{AirGlove}, an adversarial learning framework designed to learn appearance-invariant pose representations by leveraging data from existing gloves. AirGlove generalizes effectively to new glove designs, even with limited or no training data. 
Experiments with multiple sensing gloves show that AirGlove effectively generalizes the hand pose models to new glove designs and achieves a significant performance boost over the compared schemes.

\end{abstract}

\section{Introduction}
Sensing gloves have recently become essential in robotics research, supporting diverse applications such as teleoperation \cite{becker2024integratingevaluatingvisuotactilesensing, caeiro2021systematic} and human-to-robot imitation learning \cite{liu2024masked, rueckert2015lowcostsensorgloveforce}.
% In particular, sensing gloves refer to digital glove devices equipped with multiple sensors that capture both motion and tactile information from hand activities \cite{}. 
To track human hands with gloves, common approaches mainly rely on on-glove sensors like inertial measurement units (IMUs) \cite{tashakori2024capturing}. However, these solutions often suffer from instability in real-world applications as sensors often require sophisticated calibration processes and can degrade in quality over time \cite{filipowska2024machine}. Recently, significant progress has been made in vision-based pose tracking by leveraging large-scale human hand pose datasets \cite{ohkawa2023assemblyhands} and advanced models trained on diverse hand data \cite{pavlakos2023reconstructinghands3dtransformers, prakash2024wildhands}. 
With egocentric cameras, the vision-based models can provide stable and fine-grained hand pose estimation that accurately derives finger joints and wrist locations even under challenging conditions like in occlusion or with skin appearance variations~\cite{pavlakos2023reconstructinghands3dtransformers, prakash2024wildhands, cheng2024handdiff3dhandpose}.
While such models may be applied for sensing gloves, it remains unknown whether and how much the appearance discrepancies between human hands and gloves can lead to the performance degradation.
% These glove-based systems offer haptic feedback and integrate sensors directly into the glove design.
% While several powerful hand pose estimation models (e.g. HaMeR \cite{pavlakos2024reconstructing}) have emerged recently \cite{hampali2020honnotate}, their generalizability has primarily focused on bare hands that exhibit relatively uniform appearance. Gloves introduce substantial variations in shape, texture, and morphology due to task-specific designs \cite{boutis2022mchands1mglovewearinghanddataset}, which lead to significant deviations from bare-hand pose representations, degrading pose estimation performance. 
Therefore, in this work, we investigate the potential domain discrepancy between human hands and sensing gloves in the appearance-level, which is relatively unexplored despite its critical impact on glove tracking. 
% Specifically, we aim to quantitatively evaluate how glove appearance affects the performance of state-of-the-art hand tracking models and to investigate adversarial learning as a principled approach for mitigating such gaps. 

% Figure~\ref{fig:gloves} shows various glove designs that highlight the diversity in glove appearance and functionality, including gloves designed for virtual reality (VR) tracking, tactile pressure sensing, robotics training  \cite{mannam2024designinganthropomorphicsofthands}, and haptic feedback  \cite{achenbach2023give, caeiro2021systematic}. 

\begin{figure}[t]
    \centering
\includegraphics[width=0.8\linewidth]{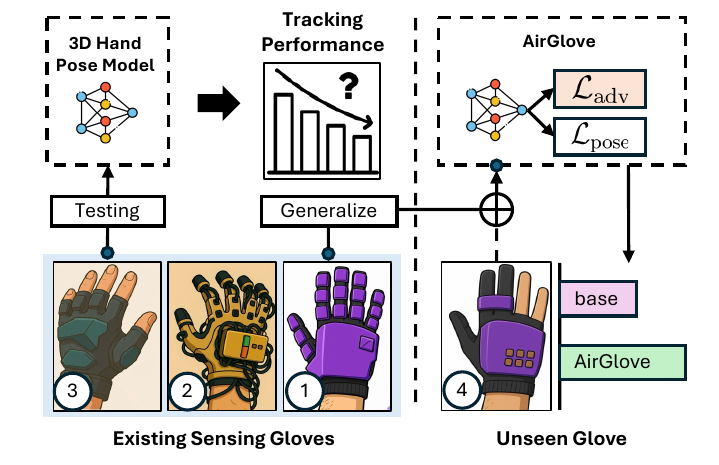}
    \caption{\emph{Overview of Our Study}. \textbf{(Left)} We quantitatively explore the potential degradation of vision-based hand tracking models on sensing gloves. \textbf{(Right)} We propose an appearance-invariant representation learning framework for glove generalization, which leverages adversarial learning on existing sensing glove data to enhance gloved hand tracking performance on unseen glove designs.}
    % Conceptual illustration of the challenges introduced by new glove appearances and the limitations of existing strategies. Our proposed AirGlove framework mitigates these issues by learning appearance-invariant pose representations, enabling effective pose generalization to unseen gloves. (The glove images are AI-generated and serve as conceptual representations for illustrative purposes. 
    \label{fig:gloves}
\end{figure}

We further illustrate the studied problem in Fig. \ref{fig:gloves}.
Sensing gloves encompass a wide variety of designs tailored to different applications.
% For instance, data-sensing gloves provide accurate motion capture for hand kinematics, while haptic gloves incorporate force-feedback mechanisms to convey tactile sensations during VR or robotic interaction. 
% Recently, \cite{kiefer2024enhanced} investigated the domain gap between human hands and sensing gloves, highlighting how visual and structural discrepancies can significantly impair the transferability of vision-based hand tracking models. 
One possible solution of adapting vision-based hand models to gloves is to collect glove-specific datasets with annotated hand poses \cite{nagymate2018application,moon2020interhand26mdatasetbaseline3d}, and then finetune the models with the data. However, the correlation between the model improvement and the data scale of sensing gloves is unclear due to the lack of sufficient glove data. Moreover, the data collection is time-consuming or even impractical in the long run given the fact that new glove designs with unique appearances are emerging. 

Motivated by these limitations, we conduct a systematic study on the performance of vision-based hand tracking models across diverse sensing gloves.
To mitigate data collection efforts for new glove designs, we introduce \textbf{AirGlove}, an \textbf{A}ppearance-\textbf{I}nvariant \textbf{R}epresentation Learning framework that leverages existing sensing gloves to learn appearance-invariant glove representations and generalizes to new gloves.
However, two technical challenges exist to achieve the above goals. The first is \emph{insufficient glove data} as there is currently no large-scale and real-world glove datasets that cover various glove applications. The second is \emph{the lack of effective design for the adversarial learning method}. While the classical adversarial framework \cite{goodfellow2014generativeadversarialnetworks} can be applied, this optimization involves two competing objectives that are inherently unstable and prone to convergence issues~\cite{wiatrak2020stabilizinggenerativeadversarialnetworks, wang2023exploringgradientexplosiongenerative}.

% To this end, we introduce \textbf{AirGlove}, an \textbf{A}ppearance-\textbf{I}nvariant \textbf{R}epresentation Learning framework designed to learn pose representations that generalize effectively to unseen glove designs. To the best of our knowledge, this is the first work to explicitly investigate and address the appearance discrepancies between sensing gloves and human hands as a domain gap for vision-based hand tracking models.
% Experimental results demonstrate that AirGlove achieves a substantial performance boost over the baseline models, with improvements up to $20.5\%$ in zero-shot evaluation and $13.2\%$ in few-shot evaluation. 
% The results provided solid proof for the outstanding capability of AirGlove in handling unseen gloves with appearance discrepancy and improved generalizability for hand pose estimation in unseen domains. 

To address the first challenge, we create a multi-sensing glove dataset with millions of video frames and Optitrack-based 3D pose labels \cite{nagymate2018application} to adapt and evaluate vision-based models.
To address the second challenge, we design an energy-based adversarial learning strategy for AirGlove by equalizing the probability of hidden glove representations across all glove appearances, thus encouraging the model to learn appearance-agnostic representations while maintaining robust pose estimation performance. 
To the best of our knowledge, our work is the first to explicitly investigate and mitigate the visual appearance discrepancies between sensing gloves and human hands for vision-based hand tracking models.
Experiments with multiple sensing gloves demonstrate substantial performance degradation of vision-based models and the significant enhancement of them by AirGlove.

\newtheorem{myDef}{Definition}
\section{Problem Definition}
% \begin{myDef}
\textbf{Definition 1. Vision-Based Gloved Hand Tracking:}
Considering a user wearing a sensing glove on their hands and an egocentric vision camera on their heads (e.g., from head-worn smart glasses or head-mount display), the user then performs various hand activities. 
We denote the recorded videos as $\mathcal{X}_g = \{x_{g,t}\}_{t=1}^T$ where $x_{g,t}$ represents $t^{\text{th}}$ frame and $g$ the specific sensing glove.
The goal of a vision-based hand tracking model is to take $x_{g,t}$ as input and predict the hand skeleton as a set of 3D landmarks. 
In particular, we denote the 21 predicted landmarks as $p_{g,t} = \{p_{g,t}^k\}_{k=1}^{21}$,
where each $p_{g,t}^k \in \mathbb{R}^3$ represents a 3D coordinate vector.
% We assume the availability of a single-view egocentric camera that records video at a fixed frame rate, capturing gloved human hands performing various hand poses. We denote the recorded video as $\mathcal{X} = \{x_t\}_{t=1}^T$, where $x_t$ represents the video frame at time step $t$, and $T$ is the total number of frames in one recording.

% \end{myDef}

% \begin{myDef}
\noindent
\textbf{Definition 2. Hand Pose Ground-Truth:} similar to model predictions, we represent the hand pose labels as a set of 3D landmarks \cite{10.1145/3386569.3392452}. Formally, the hand pose at time $t$ and for glove $g$ is denoted as
$q_{g,t} = \{q_k\}_{k=1}^{21}$. We illustrate more details of the ground-truth collection in Section \ref{cv_only}.
% \end{myDef}

% \begin{myDef}
\noindent
\textbf{Definition 3. Glove Appearance Generalization:} Given a new sensing glove with limited pose labels, our goal is to enhance vision-based models that learn appearance-invariant representations from existing sensing gloves so that the representations can be generalized to new gloves with unseen appearance. Specifically, we formulate the problem as an adversarial optimization task where an adversarial classifier mitigates the appearance gaps of different gloves in the latent features, which can be mathematically denoted as:
\begin{equation}
\mathcal{M}^* = \arg\min_{\mathcal{M}} \; \mathbb{E}_{(x, p) \sim \mathcal{X}} \left[ \mathcal{L}_{\text{pose}}(\mathcal{M}(x), p) + \mathcal{L}_{\text{adv}} \right],
\end{equation}
% \end{myDef}
where $\mathcal{M}$ denotes the vision-based model. More detailed explanation about the terms can be found in Section 4.1 and 4.2.

\begin{table}[t]

\begin{center}
\footnotesize
\resizebox{\columnwidth}{!}{
\begin{tabular}{l|c|c|c|c}
\toprule
Dataset & \# Sessions (Train) & \# Sessions (Eval) & \# Frames (Train) & \# Frames (Eval) \\
\midrule
IMU-Glove     & 28 & 6 & 858{,}059 & 446{,}200 \\
PS-Glove      & 4  & 1 & 98{,}962  & 49{,}673  \\
MoCap-Glove   & 13 & 3 & 224{,}875 & 160{,}561 \\
Haptic-Glove  & 5  & 1 & 247{,}479 & 158{,}946 \\
Bare-hand     & 5  & 3 & 232{,}104  & 56{,}883 \\
\bottomrule
\end{tabular}
}
\end{center}
\caption{Summary of Multi. Sensing Glove Dataset.}
\label{tab:glove_data}
\end{table}

\section{Vision-Based Glove Tracking}
\label{cv_only}
In this section, we detail the setup for evaluating vision-based hand tracking models on sensing gloves.
% \footnote{The actual glove appearances will be released upon paper acceptance.}

\textbf{Multi. Sensing Glove Dataset.} We adopt four representative types of sensing gloves to cover diverse use cases across major AR/VR and robotics applications. The prototypes of the gloves are shown in Fig. \ref{fig:gloves}. These include \scalebox{0.9}{\textcircled{1}} \textit{IMU-Glove} equipped with inertial sensors for capturing motion dynamics, \scalebox{0.9}{\textcircled{2}}  \textit{Haptic-Glove} integrating actuators such as motors or vibrotactile units to provide force and haptic feedback, \scalebox{0.9}{\textcircled{3}}  \textit{Pressure-sensing (PS)-Glove} with distributed pressure sensors for measuring contact interactions, and \scalebox{0.9}{\textcircled{4}}  \textit{MoCap-Glove} with optical markers for precise motion capture and annotation. 
We used Quest 3 to capture multi-view egocentric videos at 60\,Hz and an optical motion capture system to obtain ground-truth hand poses \cite{nagymate2018application}. 
% Specifically, 
% we employ UmeTrack's hand skeleton model \cite{Han_2022} for pose representation. 
% The GT skeleton consists of 20 hand joints and 1 wrist joint. 
% Each joint is defined by its position and rotation axis, while the global transformation of the hand is represented by a root transformation with 6 degrees of freedom (DoFs). 
In addition, we collect a bare-hand dataset to establish a baseline for the hand tracking models of interest. 
Each participant contributes one session, during which they are instructed to perform a variety of hand motions (e.g., swipe, pinch, free-form). For each glove, we randomly split the train/test data at the session level and show the summary of the data in Table \ref{tab:glove_data}.

\textbf{Vision-Based Bare-Hand Models.}
We adopt two vision-based bare-hand models that are directly compatible with our motion capture system annotations:
% It is trained with losses on pose, temporal consistency, and pinch detection, enabling high accuracy and robustness in VR motion and interaction tasks. 
% incorporates modules for feature unprojection, multi-view feature fusion, and temporal smoothing to handle occlusions and motion consistency. UmeTrack 
(i) \textbf{MEgATrack} \cite{10.1145/3386569.3392452}: a real-time egocentric hand-tracking framework that consists of a ResNet-based pose regressor to regress 3D hand keypoints. \cite{ohkawa2023assemblyhands}.
(ii) \textbf{UmeTrack} \cite{Han_2022}: a multi-view hand tracking framework designed for VR that takes egocentric wide-FOV camera inputs and predicts 3D hand pose in world coordinates.  Both are trained on over two million bare-hand egocentric visual data and generalize well across hand sizes and skin tones.

\begin{figure}[t]
    \centering
\includegraphics[width=0.9\linewidth]{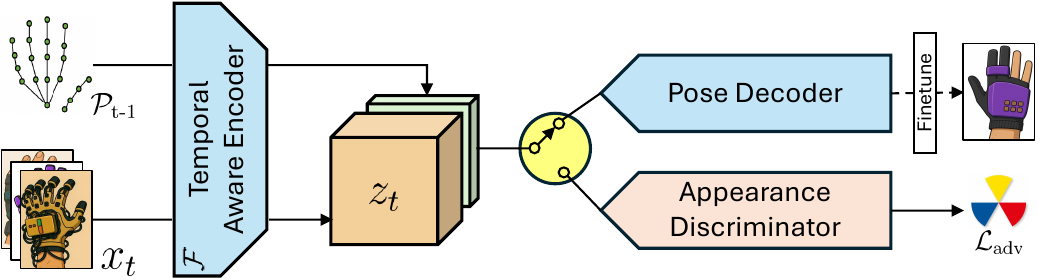}
        \caption{\emph{Overview of AirGlove.} The temporal-aware encoder extracts visual representations from egocentric videos, followed by the 3D pose decoder for pose estimation. The adversarial appearance discriminator iteratively regulates the glove representations to derive appearance-invariant features.}
        \label{fig:method}
\end{figure}

\section{Glove Appearance Generalization}
Given the exploration in Section~\ref{cv_only} on vision-based hand tracking models, we further enhance their glove tracking performance when limited annotated data is available. A simple solution is to fine-tune existing bare-hand models on the new glove data. However, when the available data scale is small, the models risk overfitting and fail to capture sufficient hand pose diversity. In addition, continuously collecting annotated datasets for every new glove design is impractical, as glove designs evolve rapidly with varying visual appearances, and obtaining 3D pose annotations is a resource-intensive process.

We propose AirGlove 
% an appearance-invariant representation learning framework that leverages the data from existing gloves with various appearances and designs an adversarial optimization strategy to jointly boost the tracking performance on new target gloves. In particular, AirGlove 
that consists of two modules: i) a temporal-aware deep visual network and ii) an adversarial appearance-invariant discriminator. In particular, AirGlove is optimized on data collected from existing gloves, denoted as $\mathcal{G}=\{g_k\}$, and expected to generalize to a new glove, denoted as $\hat{g}$. We show the overview of our proposed AirGlove framework in Fig. \ref{fig:method} and introduce each module below.

% In particular, AirGlove consists of three components: i) a temporal-aware 3D glove representation encoder (\textit{Encoder}), ii) an energy-based adversarial decoder (\textit{Decoder}), and iii) an alternating optimization strategy (\textit{Optimizer}). We introduce each component with more details below. 
% In this work, we aim to tackle the domain generalization challenge posed by distinct glove appearances when only a small number number of training samples available. We propose AirGlove to learn appearance independent features for glove pose estimation under domain gaps.

% Given a video $\mathcal{X}$, the \textit{Encoder} module aims to extract representative glove information from $\mathcal{X}$ and encode it to high-dimensional feature representations. 
% One possible solution is to encode video frames using deep computer vision models \cite{he2015deepresiduallearningimage, dosovitskiy2021imageworth16x16words}.
% However, such image-based solutions suffer from temporal inconsistency that introduces high jitter, especially for gloves with highly variable appearances \cite{khaleghi2021multiviewvideobased3dhand}.
% Therefore, it is necessary to consider historical gloved hand pose activities as context to improve glove representations.

\subsection{Temporal-Aware Deep Visual Network (TADV-Net)}
We follow \cite{10.1145/3386569.3392452} to build our visual encoder by leveraging the video frame $x_{g,t}$ and the hand pose $q_{g,t-1}$ in the previous timestamp to provide contextual motion priors.
% but also keep the information in low dimensions (i.e., 21) for efficient computation, which is important for the practical usage of the model in real-world applications \cite{chung2018hand}. 
We formulate the encoding process as $z_{g,t} = \mathcal{F}_{\text{enc}}(x_t, q_{g,t-1})$, where $z_{g,t}\in \mathbb{R}^{d}$ denotes the encoded $d$-dimensional representations. Note that during the inference stage, we replace $q_{g,t-1}$ with the $p_{g,t-1}$ to keep the auto-regressive predictions by the model.

The pose decoder is a multi-stacked MLP that converts $z_{g,t}$ to both 2D and 1D heatmaps for hand keypoints and depth, respectively.
% The pose decoder module consists of two different components: i) the heatmap-based 3D pose decoder and ii) the adversarial glove appearance discriminator. In particular, we follow \cite{10.1145/3386569.3392452} to build the 3D pose decoder by converting the learned representations $z_t$ to 2D and 1D heatmaps, where the number of each type of heatmaps equals the number of keypoints. Each 2D heatmap estimates the spatial location of a keypoint, and each 1D heatmap estimates its depth levels. 
Then a 3D hand pose loss is applied as
\begin{equation}
\mathcal{L}_{\text{pose}} = \frac{1}{K}\sum_{k=1}^{K}\|\hat{h}^\text{2D}_k - h^\text{2D}_k \|_2^2 + \|  \hat{h}^{\text{1D}}_k - h^{\text{1D}}_k\|_2^2
\end{equation}
where $K$ is the total number of keypoints, $h^{\text{1D}}$/$h^{\text{2D}}$ represent the 1D/2D heatmaps. $\hat{h}_k$ denotes the predictions.

% We adopt the model proposed in  as our baseline for performing 3D hand pose estimation from monocular video recordings. This baseline employs a variant of the ResNet34 architecture to predict 21 keypoints representing the human hand skeleton.

% Specifically, the baseline model takes as input: (1) a cropped monochrome image of the hand, obtained from bounding boxes provided by a detection network, and (2) keypoint coordinates from the previous frame, including normalized 2D coordinates and their 1D relative distance to the camera center. The model outputs two types of heatmaps for each keypoint: (1) a 2D spatial heatmap that represents the predicted locations of keypoints in the image plane, and (2) a 1D heatmap encoding the relative distance of keypoints along the depth dimension.
% Formally, this prediction process can be expressed as:
% \begin{equation}
% \mathcal{H}^{\text{2D}}[\hat{\mathcal{P}}_t],\, \mathcal{H}^{\text{1D}}[\hat{\mathcal{P}}_t] = \mathcal{M}(x_t, \mathcal{P}_{t-1}),
% \end{equation}
% where $\mathcal{H}[\hat{\mathcal{P}}_t^{\text{2D}}]$ and $\mathcal{H}[\hat{\mathcal{P}}_t^{\text{1D}}]$ denote the predicted 2D spatial and 1D relative distance heatmaps, respectively. 
%  An illustration of the baseline model is provided in Figure~\ref{fig:method}.
% This dual-output design leverages both spatial and geometric information, enhancing pose estimation accuracy.

\begin{table}[t]

\begin{center}

\footnotesize
\resizebox{\columnwidth}{!}{
\begin{tabular}{l|l|c|c|c|c}
\toprule
Eval Data & Model & MKPE ($\downarrow$) & MKPE.T ($\downarrow$) & F-MKPE ($\downarrow$) & F-MKPE.T ($\downarrow$) \\
\midrule

\multirow{2}{*}{Bare-hand} 
 & MEgATrack & 16.677 & 8.408 & 21.162 & 16.325 \\
 & UmeTrack  & 15.150 & 7.970 & 18.574 & 15.142 \\
\midrule

\multirow{2}{*}{IMU-Glove} 
 & MEgATrack & 33.938 & 13.009 & 40.049 & 25.837 \\
 & UmeTrack  & 18.559 & 10.273 & 22.823 & 20.286 \\
\midrule

\multirow{2}{*}{PS-Glove} 
 & MEgATrack & 73.979 & 19.401 & 78.751 & 39.514 \\
 & UmeTrack  & 54.601 & 13.804 & 62.398 & 27.350 \\
\midrule

\multirow{2}{*}{MoCap-Glove} 
 & MEgATrack & 21.487 & 11.976 & 27.046 & 22.967 \\
 & UmeTrack  & 57.452 & 12.099 & 62.953 & 23.146 \\
\midrule

\multirow{2}{*}{Haptic-Glove} 
 & MEgATrack & 63.509 & 16.232 & 66.466 & 32.151 \\
 & UmeTrack  & 60.525 & 14.840 & 66.473 & 29.190 \\
\bottomrule

\end{tabular}
}
\end{center}
\caption{\emph{Comparison of pose tracking results (mm) on bare-hand and sensing gloves.} Both evaluated models show substantial performance degradation when applied to gloves.}

\label{tab:zero_shot_results}
\end{table}
\subsection{Adversarial Appearance-Invariant Discriminator}
% The goal of the \textit{Decoder} module is to derive 3D glove poses from the  representation learned by the \textit{Encoder} while suppressing the glove appearance-specific information from the representations.
% As indicated in Section 1, directly fusing data from all existing gloves may not help or even degrade the model performance. Therefore, we propose an energy-based adversarial decoder that modulates the learned representations from the existing gloves. 

% Then a 3D hand pose loss is applied as
% \begin{equation}
% \mathcal{L}_{\text{pose}} = \frac{1}{K}\sum_{k=1}^{K}\|\hat{h}^\text{2D}_k - h^\text{2D}_k \|_2^2 + \|  \hat{h}^{\text{1D}}_k - h^{\text{1D}}_k\|_2^2
% \end{equation}
% where $K$ is the total number of keypoints, $h^{\text{1D}}$/$h^{\text{2D}}$ represent the 1D/2D heatmaps. $\hat{h}_k$ represents the predicted heatmaps by the pose decoder for keypoint $k$.

The design of the Adversarial Appearance-Invariant Discriminator (\textbf{AAID}) is based on the assumption that each glove with a unique appearance can be considered as a semantic class. Given the representation $z_{g,t}$ of a particular frame, AAID should assign equal probabilities to all gloves if $z_{g,t}$ is not biased to any glove appearance. Specifically,
we construct a convolutional classifier $\theta_{\text{cls}}$ that takes $z_{g,t}$ as input and outputs $|\mathcal{G}|$ dimensions as predicted classes. To suppress appearance information, 
we define the adversarial loss $\mathcal{L}_{\text{adv}}$ as the Kullback–Leibler (KL) \cite{kullback1951information} divergence between the classifier output and a uniform prior $U$. Minimizing the divergence can be formulated as $\mathcal{L}_{\text{adv}}$:
\begin{equation}
D_{\text{KL}}(U\,\|\,\theta_{\text{cls}}(z) )= -\frac{1}{|\mathcal{G}|} \sum_{c=1}^{|\mathcal{G}|} \log \theta_{\text{cls}}(z) - \log |\mathcal{G}|
\end{equation}
where $\theta_{\text{cls}}(z)$ is the output of the classifier. The total optimization objective for AirGlove is defined as $\mathcal{L}_{\text{total}} = \mathcal{L}_{\text{pose}} + \lambda_{\text{adv}} \mathcal{L}_{\text{adv}}$ where $\lambda_{\text{adv}}\in [0, 1]$.

We adopt an alternating optimization strategy to jointly optimize pose estimation and disentangle glove-specific appearances. The process alternates between two phases: (i) freeze TADV-Net and update \(\theta_{\text{cls}}\) by minimizing the multiclass cross-entropy \(\mathcal{L}_{\text{cls}}\);
(ii) freeze \(\theta_{\text{cls}}\) and update TADV-Net by minimizing \(\mathcal{L}_{\text{total}}\). These two phases alternate every $E=3$ epochs. 
By ensuring that AAID adapts to the evolving visual representations while TADV-Net improves its ability to challenge the discriminator in return, AirGlove effectively learns appearance-invariant glove representations.

\section{Evaluation}
We conduct extensive experiments on the collected datasets to answer the following research questions:
\textbf{Q1}) How much does the pose tracking performance degrade when applying the vision-based bare-hand models directly to sensing gloves? 
\textbf{Q2}) Given a new glove with limited annotated data, how effectively can AirGlove enhance the glove tracking performance?  
\textbf{Q3}) How well does AirGlove learn appearance-invariant representations from existing sensing gloves?

\begin{figure}[t]
  \centering
    \includegraphics[width=\linewidth]{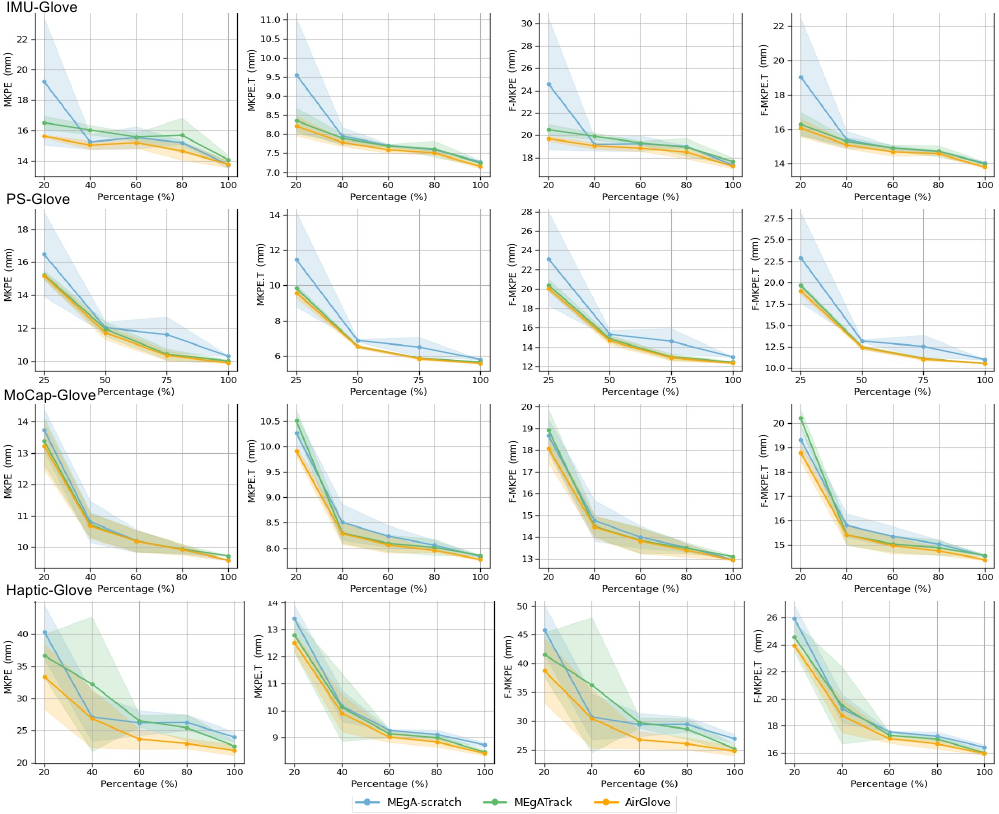}
 \caption{Evaluation of AirGlove on sensing glove datasets. AirGlove achieves superior tracking performance compared to all baselines across different sensing gloves (row-wise) and evaluation metrics (column-wise). Best viewed in color.
}
  \label{fig:few-shot}
\end{figure}

\textbf{Evaluating Bare-hand Models on Gloves (Q1).}
% Given an unseen glove with no available data, we trained the baseline pose estimation model \textbf{MEgA-src} and AirGlove on the source glove datasets $\mathcal{D}_{\text{src}}$, respectively. The trained models were then evaluated on the evaluation set of the unseen glove dataset $\mathcal{D}_{\text{tgt}}$. The experimental results are summarized in Table~\ref{tab:zero_shot_results}. Our results show that AirGlove consistently outperforms MEgA-src across all gloves and evaluation metrics. Notably, AirGlove achieves a significant MKPE reduction of $\mathbf{20.5\%}$ compared to MEgA-src for Glove-D. We attribute this substantial improvement to the \textit{Decoder} module of AirGlove, which effectively learns appearance-invariant pose representations to enhance generalization performance.
We follow \cite{10.1145/3386569.3392452} to adopt \textit{Mean Keypoint Position Error} (MKPE) and Fingertip MKPE (F-MKPE), together with their transformed variants (MKPE.T and F-MKPE.T) as the evaluation metrics. We show the evaluation results in  Table \ref{tab:zero_shot_results}. 
We observe that both MEgATrack and UmeTrack show substantially degraded performance on sensing gloves compared to bare-hands. 
% For instance, PS-Glove and Haptic-Glove exhibit the largest errors due to the large appearance variations introduced by sensors on the glove. 
The findings highlight the significant gap between sensing gloves and bare-hands in vision-based methods, emphasizing the challenge with appearance discrepancies. 

\textbf{Evaluating AirGlove in Fine-tuning Settings (Q2).}
To answer Q2, we take MEgATrack as the baseline to compare with AirGlove as they share the same visual encoder backbone. We train both MEgATrack and AirGlove on the union of three sensing glove datasets, leaving one glove out as the held-out set. 
% This setup simulates the practical scenario of evaluating a new glove unseen during training. 
% All models are trained on the training sessions of the Multi-Glove dataset and evaluated on the eval sessions. 
Given a new target glove, 
we fine‐tune both MEgATrack and AirGlove on increasing proportions of the training data from 20\% to 100\%. For each proportion, we randomly sample the data 5 times and report the averaged results with standard deviation. We also propose \textbf{MEgA-scratch} that is trained from scratch on the same data proportion of the target glove. We show the quantitative evaluation results in Fig. \ref{fig:few-shot}. We observe that AirGlove consistently outperforms all compared schemes at every data proportion and across all metrics.
We attribute the generalization performance of AirGlove to its adversarial learning strategy, which effectively learns pose representations that are invariant to appearances. 
% Notably, when Glove-D is held out, AirGlove fine-tuned with only 60\% of the data achieves MKPE performance equivalent to MEgA-scratch trained on the full 100\% dataset. Moreover, for any proportion up to 80\%, AirGlove reduces MKPE and F-MKPE by more than $10\%$ relative to MEgA-src. Considering our pilot study in Section 1, which demonstrated that naively training on the union of source and target glove datasets yields suboptimal performance, 

\begin{figure}[t]
    \centering
    \includegraphics[width=0.8\linewidth]{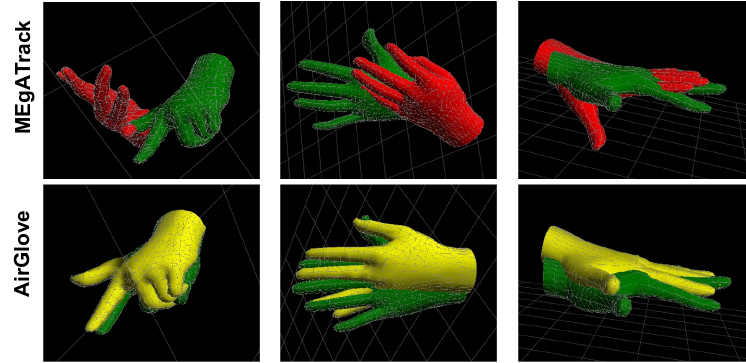}
    \caption{\emph{Visualization for AirGlove.} Compared to the baseline MEgATrack (red), AirGlove (yellow) generates predictions that are better aligned with the ground-truth (green).}
    \label{fig:hand-vis}
\end{figure}

We qualitatively evaluate the tracking performance of AirGlove by rendering 3D hand meshes based on the predicted hand poses \cite{salter2024emg2poselargediversebenchmark}. For Haptic-Glove as the unseen glove, we compare reconstructed meshes from AirGlove and MEgATrack. As shown in Fig.~\ref{fig:hand-vis}, AirGlove yields hand poses more closely aligned with ground truth than the baseline.

\begin{figure}[t]
    \centering
\includegraphics[width=0.9\linewidth]{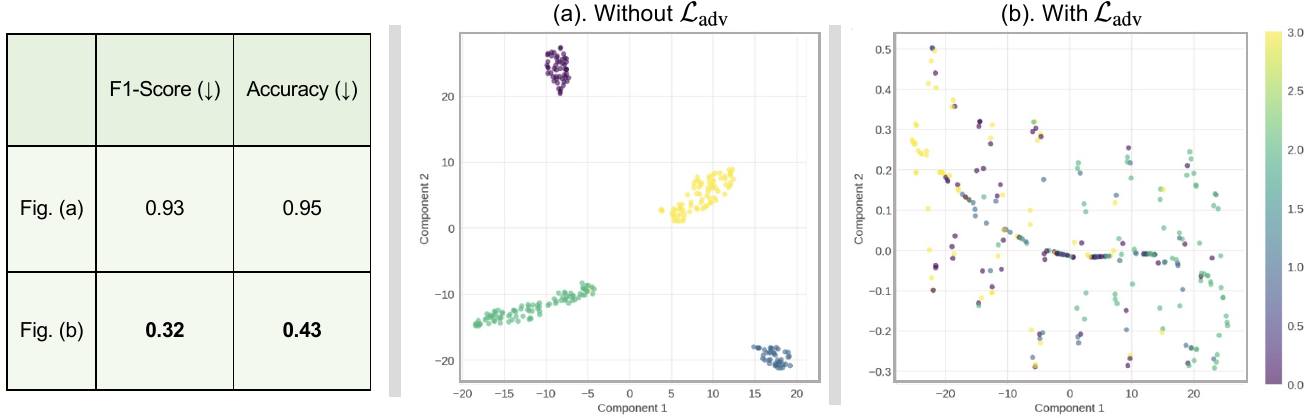}
    \caption{\emph{Glove classification results and t-SNE visualization of glove representations.} With $\mathcal{L}_\mathrm{adv}$ vs. without, the model learns features that cannot differentiate glove appearances, leading to pose representations without appearance bias.}
    \label{fig:features}
\end{figure}

\textbf{Ablation Study (Q3)}
We train AirGlove on the union of all glove datasets, once with \(\mathcal{L}_{\mathrm{adv}}\) enabled and once without it. After training, we freeze the pose encoder and evaluate the adversarial classifier \(\theta_{\text{cls}}\) on held-out samples to measure classification \emph{Accuracy} and \emph{F1-score} \cite{novakovic2017evaluation}. We also extract the outputs from \(\theta_{\text{cls}}\) and randomly select a subset of samples for visualization using t-SNE \cite{van2008visualizing}.
Fig.~\ref{fig:features} shows both quantitative and qualitative results. 
We observe that enabling $\mathcal{L}_{\mathrm{adv}}$ significantly degrades the classification performance, which is aligned with the visualization where the feature clusters for each glove collapses into a mixed distribution.
% Without \(\mathcal{L}_{\mathrm{adv}}\), \(\theta_{\text{cls}}\) achieves Accuracy = $0.95$ and F1-score = $0.93$, and the t-SNE plot reveals distinct clusters for each glove. With \(\mathcal{L}_{\mathrm{adv}}\), classification Accuracy falls to $0.43$ and F1-score to $0.32$, and the feature clusters collapsed into a mixed distribution. 
These results confirm that $\mathcal{L}_{\mathrm{adv}}$ effectively suppresses appearance-specific information in representations learned from the pose encoder, thus enhances generalization under glove appearance gaps.

\section{Conclusion}
In this work, we investigated the hand tracking problem for sensing gloves, where appearance variations introduced by diverse glove designs severely degrade the performance of vision-based bare-hand models. To mitigate the issue, we introduced AirGlove that learns appearance-invariant representations, enabling generalizable glove tracking without the need for large-scale training data on a new glove design. 
% Our framework integrates an energy-based adversarial loss with an alternating optimization strategy to effectively disentangle appearance from pose features while maintaining stability during training.  
% AirGlove enables effective pose estimation on new glove designs with little additional data collection. 
Experimental results demonstrate that AirGlove can enhance the generalization of gloved hand tracking on new glove designs, which consistently outperforms comparing schemes.

\vfill\pagebreak

\bibliographystyle{IEEEbib}
\bibliography{refs}

\end{document}